\documentclass{article}

\usepackage{times}
\usepackage{epsfig}
\usepackage{graphicx}
\usepackage{amsmath}
\usepackage{amssymb}
\usepackage{booktabs}
\usepackage{multirow}
\usepackage{color}
\usepackage{filecontents}
\usepackage[noadjust]{cite}
\usepackage{pgfplots,pgfplotstable,filecontents}
\usepackage{caption}
\pgfplotsset{
    compat=1.5,
    every axis/.append style={
        legend style={font=\small}}
        }
\usepackage{tabularx}

\usepackage[pagebackref=true,breaklinks=true,letterpaper=true,colorlinks,bookmarks=false]{hyperref}

\usepackage{pifont}

\def\eg{\emph{e.g}. }

\def\etal{\emph{et al}. }
\newcommand\ie{\textit{i.e. }}
\begin{document}
%
\title{Generic 3D Convolutional Fusion\\ for image restoration}
%
%
%

\author{Jiqing~Wu,
        Radu~Timofte,
        and~Luc~Van~Gool}
\maketitle

\begin{abstract}
Also recently, exciting strides forward have been made in the area of image restoration, particularly for image denoising and single image super-resolution. Deep learning techniques contributed to this significantly. The top methods differ in their formulations and assumptions, so even if their average performance may be similar, some work better on certain image types and image regions than others. This complementarity motivated us to propose a novel 3D convolutional fusion (3DCF) method. Unlike other methods adapted to different tasks, our method uses the exact same convolutional network architecture to address both image denoising and single image super-resolution. As a result, our 3DCF method achieves substantial improvements (0.1dB-0.4dB PSNR) over the state-of-the-art methods that it fuses, and this on standard benchmarks for both tasks. At the same time, the method still is computationally efficient.
\end{abstract}


\section{Introduction}
\label{sec:introduction}

Image restoration is concerned with the reconstruction/estimation of the uncorrupted image from a corrupted or incomplete one. Typical corruptions include noise, blur, down-sampling, hardware constraints (\eg Bayer pattern) and combinations of those. After decades of research there is a large literature~\cite{Katsaggelos-BOOK-2012} dedicated to restoration tasks, whereas the literature studying the fusion of restoration results is thin~\cite{Stathaki-BOOK-2011}. In this paper we tackle such fusion as a means for further performance improvements. Particularly, we propose a 3D convolutional fusion (3DCF) method and validate it on image denoising and single image super-resolution.


\subsection{Image denoising (DN)}

Natural image denoising aims at recovering the clean image given a noisy observation. The most often studied case is when the image corruption is caused by additive white Gaussian (AWG) noise with known variance.
Also, the images are assumed to be natural, capturing every-day scenes, and the quantitative measure for assessing the recovery result is the peak signal-to-noise ratio (PSNR), which stands in monotonic relation to the mean squared error (MSE).

The most successful denoising methods employ at least one of the following \noindent{\textbf{Denoising principles}} as listed in~\cite{Lebrun-AN-2012}:
Bayesian modeling (coupled with Gaussian models for noiseless patches), transform thresholding (assumes sparsity of patches in a fixed basis),
sparse coding (sparsity over a learned dictionary),
pixel or block averaging (exploits image self-similarity).

Most denoising methods work at a single image scale, the finest one, and
often a small image patch is the basic processing unit. The patch
captures local image information for a central pixel and a statistical amount of uncorrupted pixels.
Zontak~\etal~\cite{Zontak-CVPR-2013} recently opened up a fresh research direction by proposing a method based on patch recurrence across scales (PRAS).
Another partition of the methods is based on whether only the noisy image is used, or also learned priors and/or extra data from other (clean) natural images.
This leads to \textit{internal} and \textit{external} methods. Some well known examples of each are:

\noindent{\textbf{Internal denoising methods:}}

\noindent{\bf NLM} (non-local means)~\cite{buades2005non} reconstructs a noisy patch with a weighted average of similar patches from the same image.
It uses the image self-similarity and the fact that the noise is usually uncorrelated.

\noindent{\bf BM3D} (block matching 3D)~\cite{Dabov-TIP-2007} extends NLM and the DCT denoising method~\cite{Yu-IPOL-2011}.
BM3D groups similar patches into a 3D block, applies 3D linear transform thresholding, and inverses the transform.

\noindent{\bf WNNM} (weighted nuclear norm minimization)~\cite{Gu-CVPR-2014} follows the self-similarity principle,
and applies WNNM to recover the noiseless patch from a matrix of stacked non-local similar patch vectors.

\noindent{\bf PRAS} (patch recurrence across scales)~\cite{Zontak-CVPR-2013} creates (an)isotropic image scale pyramids and extracts the estimated
(noiseless) patch from the same corresponding position but at a different scale.

\noindent{\bf PLE} (piecewise linear estimation)~\cite{Yu-TIP-2012} is a Bayesian restoration model, including denoising, deblurring, and inpainting. PLE employs a set of 19 Gaussian models obtained from synthetic edge images (as priors) and an estimation-maximization iterative procedure.

\noindent{\textbf{External denoising methods:}

\noindent{\bf EPLL} (expected patch log likelihood)~\cite{Zoran-ICCV-2011} can be seen as a shotgun extended version of PLE.
It learns a Gaussian mixture model with 200 components for 2 million clean patches sampled from external natural images, and tries to maximize the expected log likelihood of any randomly chosen patch in the image.

\noindent{\bf LSSC} (learned simultaneous sparse coding)~\cite{Mairal-ICCV-2009} adapts a sparse dictionary learned over an external database by adding a grouping step to the noise image.

\noindent{\bf MLP} (multi-layer percepton)~\cite{Burger-CVPR-2012} learns from an external database with clean and noisy images, and was among the first to introduce neural networks to low level image restoration tasks.

\noindent{\bf CSF}(cascade of shrinkage fields)~\cite{schmidt2014shrinkage} proposes shrinkage fields, combining the image model and the optimization algorithm as a whole. The time complexity is greatly reduced by inherent parallelism.

\noindent{\bf opt-MRF} (Loss-Specific Training of Filter-Based MRFs)~\cite{chen2013revisiting} revisits loss-specific training and uses  bi-level optimization to solve the image restoration problem.

\noindent{\bf TRD} (trained reaction diffusion)~\cite{chen2015learning} extends the solving process of nonlinear reaction diffusion to a deep recurrent neural network, outperforms many of the aforementioned methods, 
while offering the lowest time complexity for now.
 

It is quite surprising that most of the recent top denoising methods (such as BM3D, LSSC, EPLL, PRAS, and even WNNM) face a plateau. They perform equally well for a large range of noise, despite that they are quite different in their formulations, assumptions, and information used. 
This is the reason behind the recent work that fuses them, pushing the limits by combining different approaches~\cite{Jancsary-ECCV-2012,Burger-GCPR-2013}. We refer the readers to~\cite{Stathaki-BOOK-2011} for a study of image fusion algorithms of the past decades.
Others investigated the theoretical limits for denoising with natural image patch priors~\cite{Levin-ECCV-2012}, and at least for the lower noise levels,
the gap between the most successful methods and the predicted limits seems to rapidly diminish.

\noindent{\textbf{Fusion methods:}}

\noindent{\bf PatchSNR} (patch signal-to-noise ratio). Mosseri~\etal~\cite{Mosseri-ICCP-2013}
propose a patch-wise signal-to-noise-ratio to distinguish whether an internal or an external denoising method should be applied. Their fused result slightly improves over the stand-alone methods.

\noindent{\bf RTF} (regression tree fields). Jancsary~\etal~\cite{Jancsary-ECCV-2012} observe that depending on the image content some methods perform better than other.
They consider RTFs based on a filterbank (RTF$_{plain}$), also additional exploitation of BM3D's output (RTF$_{BM3D}$),
or a setting exploiting all the outputs of their benchmarked methods (RTF$_{all}$).
The more methods the better their fusion result. The RTFs are learned on large datasets. It is also worth mentioning that following~\cite{Jancsary-ECCV-2012}, Schmidt~\etal~\cite{schmidt2014cascades} propose a cascade of regression tree fields (CRTF) working on deblurring and denoising and obtain good performances in both cases. 

\noindent{\bf NN} (neural nets / multi-layer perceptron). Burger~\etal~\cite{Burger-GCPR-2013} pursue the success of MLP~\cite{Burger-CVPR-2012} in denoising, to learn the best fusion. They found the internal denoising methods to suit better images with artificial (human-made) contents, and external ones to work better for natural scenes. They argue against PatchSNR and consider that there is no trivial rule to decide among internal or external method on a patch-by-patch basis, and indeed their NN fusion produces the best denoising results to date.
Unfortunately, the learning is quite intensive.


\subsection{Single image super-resolution (SR)}

Single image super-resolution (SR) is another active area~\cite{Timofte-arxiv-2015,kim2015deeply,wang2015deep} of image restoration aiming at recovering a high-resolution (HR) image from a low-resolution (LR) input image by inferring missing high frequency contents. We can roughly categorize the recent methods in:

\noindent{\textbf{Non-neural network methods:}}

\noindent{\bf SR} (sparse representation)~\cite{yang2010image} generates a sparse representation/coding of each LR image patch, and then applies the coefficients of this representation to generate the HR image.

\noindent{\bf A+} (adjusted anchored neighborhood regression)~\cite{timofte2014a+}, considered to be an advanced version of ANR (anchored neighborhood regression)~\cite{timofte2013anchored}, learns sparse dictionaries and regressors anchored to the dictionary atoms. 

\noindent{\bf RFL} (super-resolution Forests)~\cite{schulter2015fast} maps low to high-resolution patches using random forests and anchored regressors as in A+.

\noindent{\bf selfEx} (transformed self-exemplars)~\cite{huang2015single} introduces a self-similarity based image SR algorithm by applying transformed self-exemplars.

\noindent{\textbf{Neural network methods:}}

\noindent{\bf SRCNN} (convolutional neural network)~\cite{dong2015image} learns an end-to-end mapping between the low/high-resolution images by a deep convolutional neural network.

\noindent{\bf CSCN} (cascade of sparse coding network)~\cite{wang2015deep} combines the key ingredients of deep learning with those of the sparse coding model.


\subsection{Contributions} 

In this paper, we study the patch-by-patch fusion of image restoration methods with particular focus on recent top methods for both DN and SR tasks.
To this end, we propose a generic 3D convolutional fusion architecture (3DCF) to learn the best combination of existing methods.
Our three main contributions are:
\begin{enumerate}
\item We show the complementarity of different methods (\eg internal vs. external).
\item We demonstrate that our method learns sophisticated correlation details from top methods to achieve the best reported results on a wide range of images.
\item The generality of our 3DCF method for both DN and SR.
\end{enumerate}

The paper is organised as follows. 
Section~\ref{sec:insights} provides some insights and empirical evidence for the complementarity of the DN/SR methods and analyses oracle bounds for fusion. Section~\ref{sec:3D} motivates and introduces our novel 3DCF method with the necessary details and mathematical formulations.
Section~\ref{sec:experiments} presents the experiments and discusses the results. Section~\ref{sec:conclusions} concludes the paper.


\section{Insights}
\label{sec:insights}

Our focus is fusion for improved image restoration results and particularly for denoising in the presence of additive white Gaussian noise (AWG), with validation on single image super-resolution.
Here we analyse the complementarity of the restoration methods and fusion strategies.


\subsection{Complementarity of top methods}
\label{ssc:complementarity}

Jancsary~\etal~\cite{Jancsary-ECCV-2012}, Burger~\etal~\cite{Burger-GCPR-2013}, and Zontak and Irani~\etal~\cite{Zontak-CVPR-2013}, among others, already observed that each method works best for some particular image contents while being worse than others for other image regions.

First, we pair-wise compare the PSNR performances of BM3D (internal method), and MLP and TRD (external methods) on 68 images from the Berkeley dataset for AWG noise with $\sigma=50$. The relative improvements (PSNR gain) are reported in Fig.~\ref{fig:no_winner}. MLP is better than BM3D on all images but is worse than TRD on $\sim40\%$ of them. Also, BM3D is better than TRD on some images. We conclude there is no absolute winner at image-level.

Second, we compare pixel-wise or patch-wise and see that within the same image there is no absolute winner always getting the best result either. 
In Fig.~\ref{fig:selections} for one image altered with AWG noise, $\sigma=50$, we report pixel-wise selections from BM3D (25.77dB PSNR) and MLP (26.19dB) to best match the ground truth image. Despite MLP being significantly better (+0.41dB) on denoising this image, at pixel-level the results are almost equally divided between the methods. At patch-level (sizes $5\times5$ and $17\times17$ pixels) we have a similar pattern. 

\begin{figure}[t!]
    \centering
    \resizebox{0.8\linewidth}{!}
    {
    \setlength{\tabcolsep}{7pt}
    \begin{tabular}{ccc}
    \includegraphics[width=0.5\linewidth]{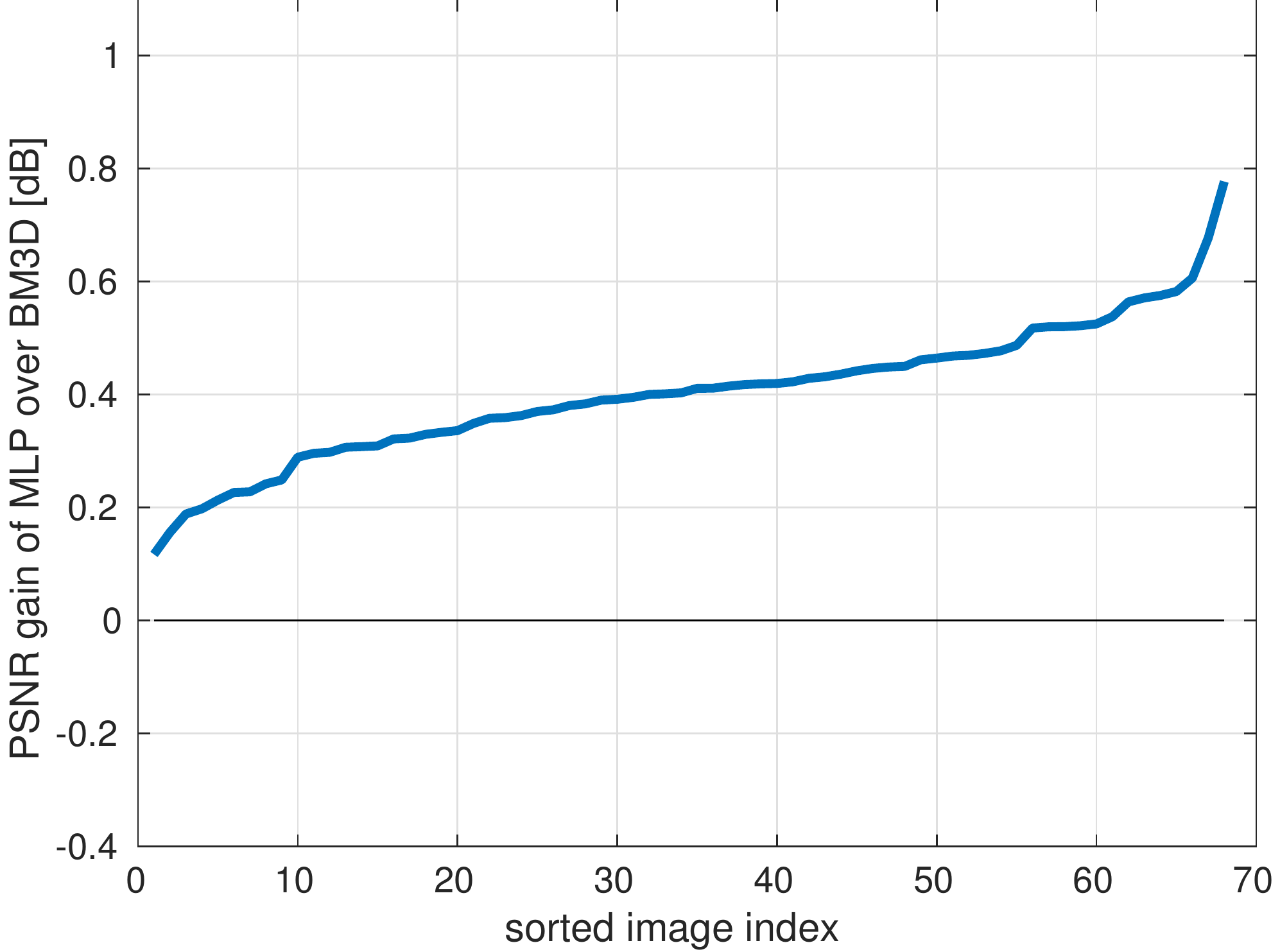}&
    \includegraphics[width=0.5\linewidth]{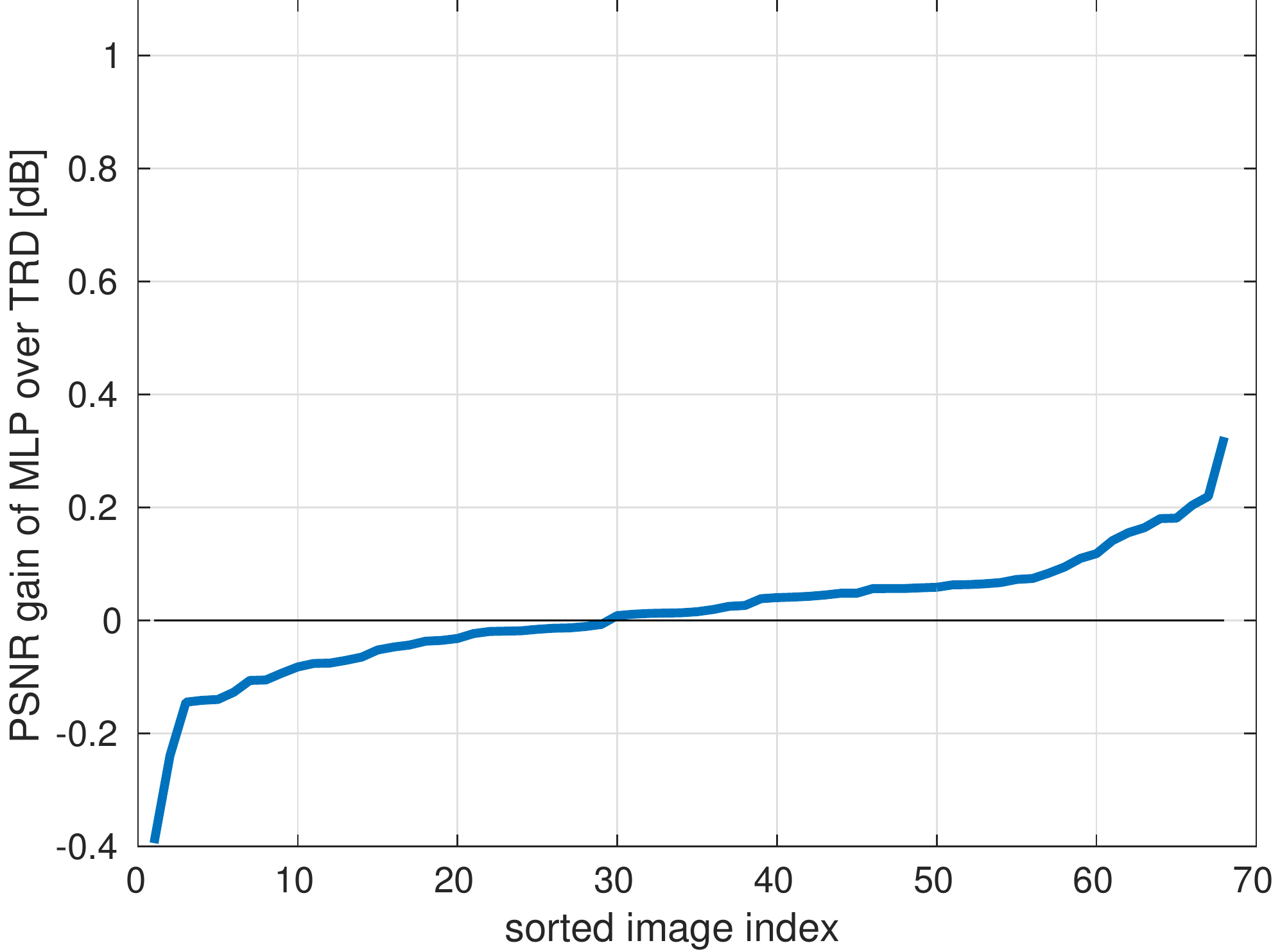}&
    \includegraphics[width=0.5\linewidth]{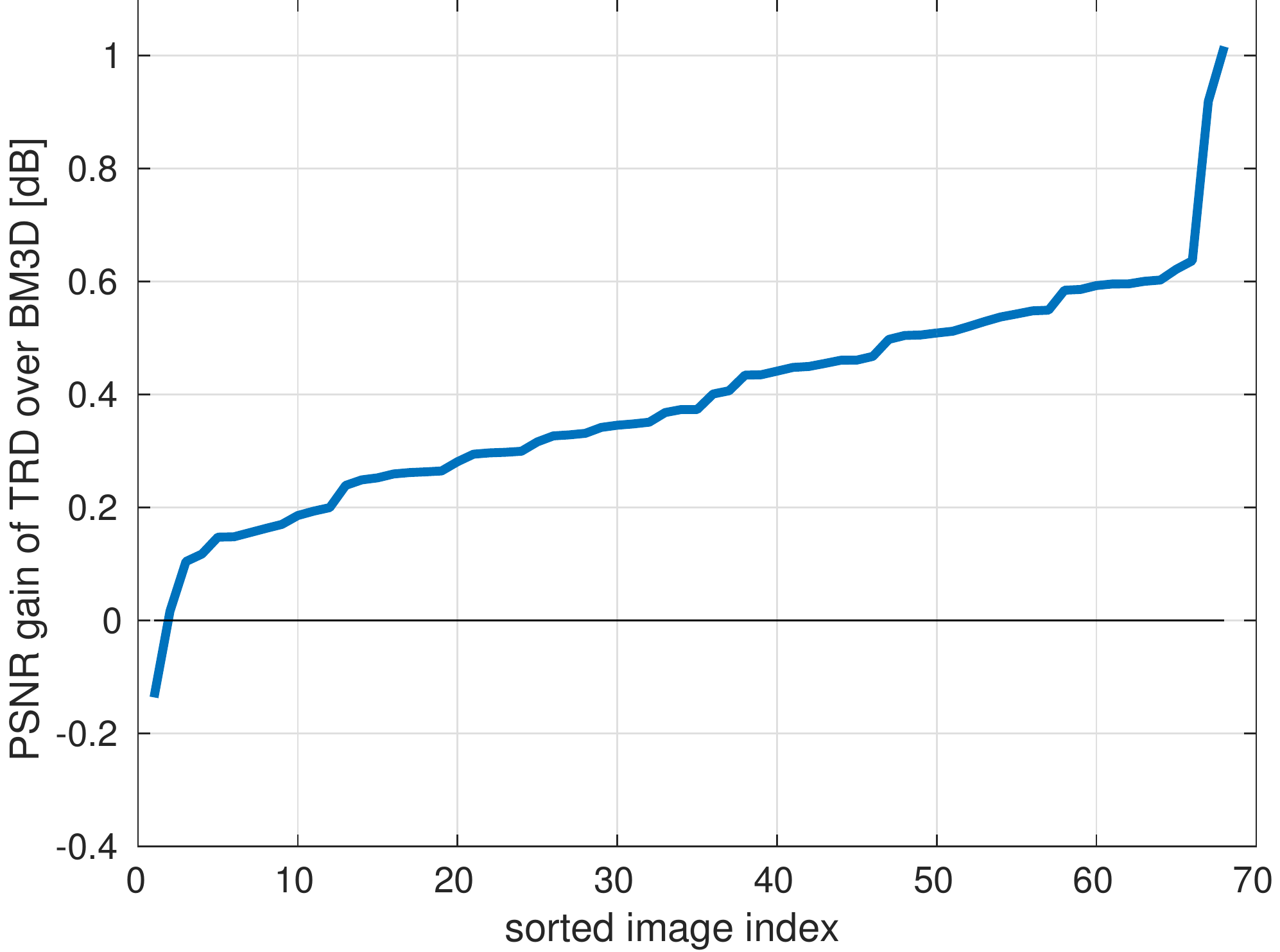}\\
    (a) MLP vs. BM3D & (b) MLP vs. TRD & (c) TRD vs. BM3D\\
    \end{tabular}
    }
    
    \caption{No absolute winner. Each method is trumped by another on some image.}
    \label{fig:no_winner}
   
\end{figure}

\begin{figure}[b!]
    \centering
    \resizebox{\linewidth}{!}
    {
    \begin{tabular}{cccc}
    \includegraphics[width=0.5\linewidth]{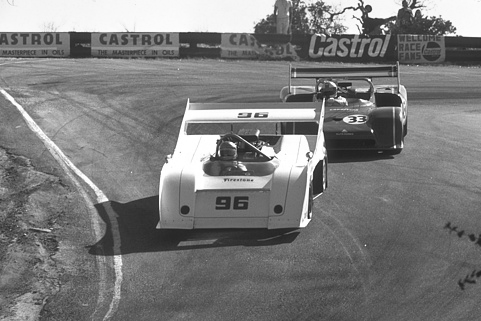}&
    \includegraphics[width=0.5\linewidth]{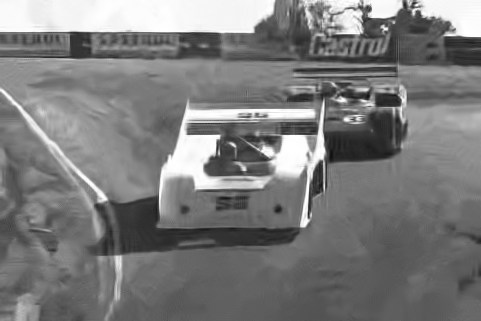}&
    \includegraphics[width=0.5\linewidth]{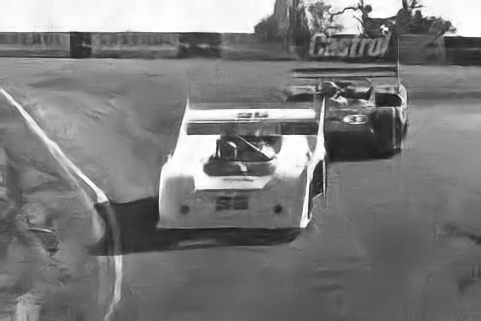}&
    \includegraphics[width=0.5\linewidth]{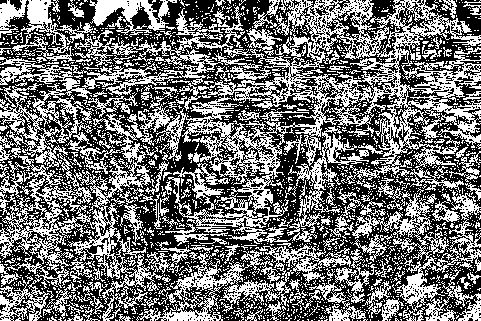}\\
    Ground truth &  BM3D (25.77dB) & MLP (26.19dB) & pixel-wise (27.01dB)\\
    \includegraphics[width=0.5\linewidth]{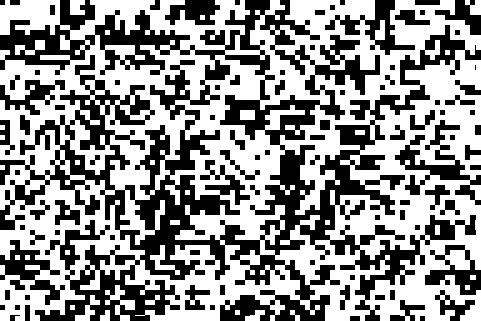}&
    \includegraphics[width=0.5\linewidth]{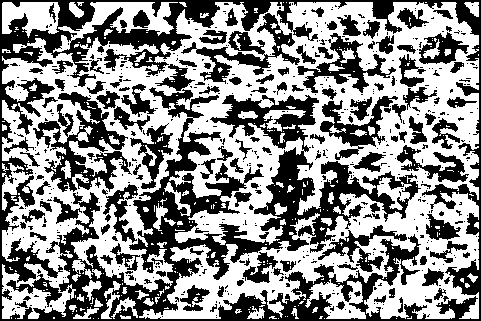}&
    \includegraphics[width=0.5\linewidth]{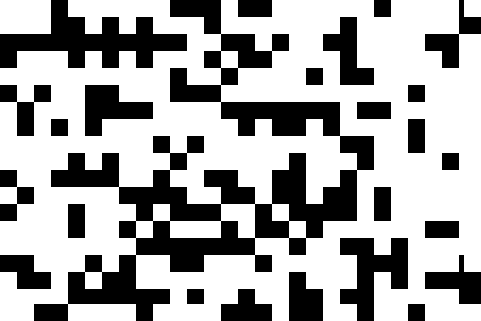}&
    \includegraphics[width=0.5\linewidth]{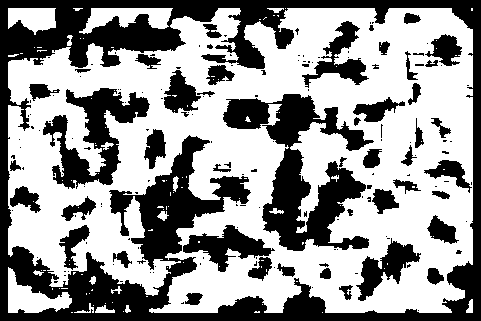}\\
    5x5 patch (26.46dB) &  5x5 patch overlapped (26.52dB) & 
    17x17 patch (26.27dB)& 17x17 patch overlapped (26.32dB)\\
    \end{tabular}
    }
    
    \caption{An example of oracle pixel and patch-wise selections from BM3D and MLP outputs and the resulting PSNRs for AWG with $\sigma=50$.}
    \label{fig:selections}
\end{figure}


\subsection{Average and selection fusion and oracle bounds}
\label{ssc:fusion_bounds}

As shown in Fig.~\ref{fig:no_winner} for images and in Fig.~\ref{fig:selections} for patch or pixel regions, the denoising methods are complementary in their performance. 
Now we study a couple of fusion strategies at image level. 

\noindent{\bf Average fusion} directly averages the image results.

\noindent{\bf Selection of non-overlapping patches} assumes that the fusion result contains non-overlapping (equal size) patches with the best image results of the fused methods (see Fig.~\ref{fig:selections}). One needs to learn a patch-wise classifier.

\noindent{\bf Selection of overlapping patches} is similar to the above one in that a patch-wise decision is made, but this time the patches overlap. The final fusion result is obtained by averaging the patches in the overlapped areas (see Fig.~\ref{fig:selections}).

We work with BM3D and MLP, partly because BM3D is an internal while MLP is an external method, and partly because of the result in Fig.~\ref{fig:no_winner} where at image level MLP performs better than BM3D. Therefore, the results from fusing BM3D and MLP at patch-level are interesting to see.

\begin{figure}[]
\centering
  \begin{minipage}[c]{0.49\textwidth}
    \centering
    \includegraphics[width=\linewidth]{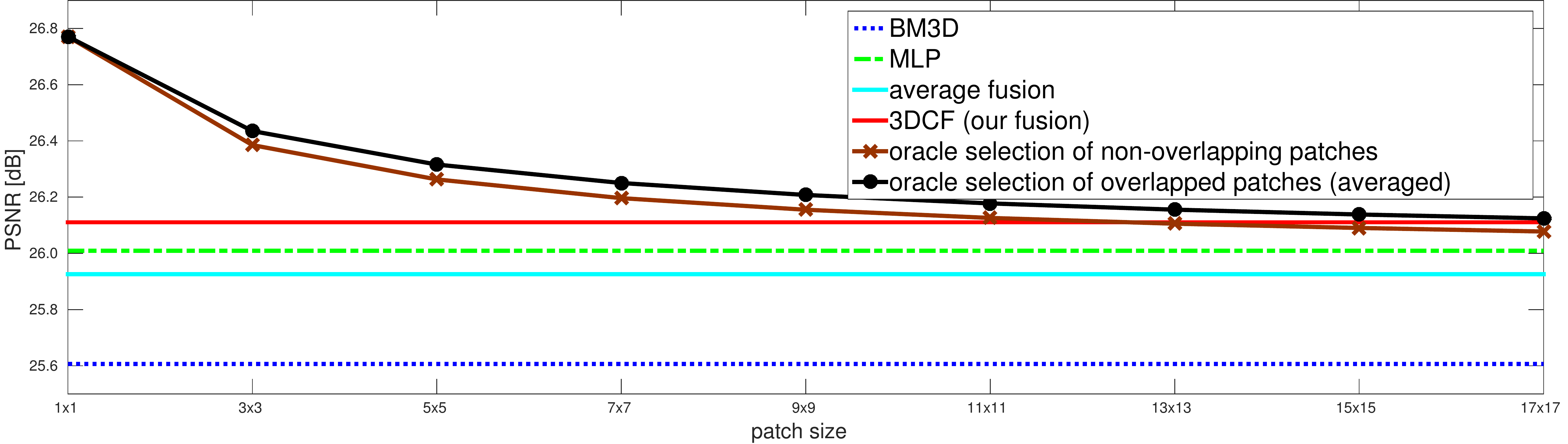}
    
    \caption{Average PSNR [dB] comparison of BM3D~\cite{Dabov-TIP-2007} and MLP~\cite{Burger-CVPR-2012}, average fusion, oracle selection of (overlapping or non-overlapping) patches, and our 3DCF fusion on 68 images, with AWG noise, $\sigma=50$.}
    \label{fig:fusion_oracle}
        \end{minipage}\hfill
  \begin{minipage}[c]{0.49\textwidth}
    \centering
    \includegraphics[width=\linewidth]{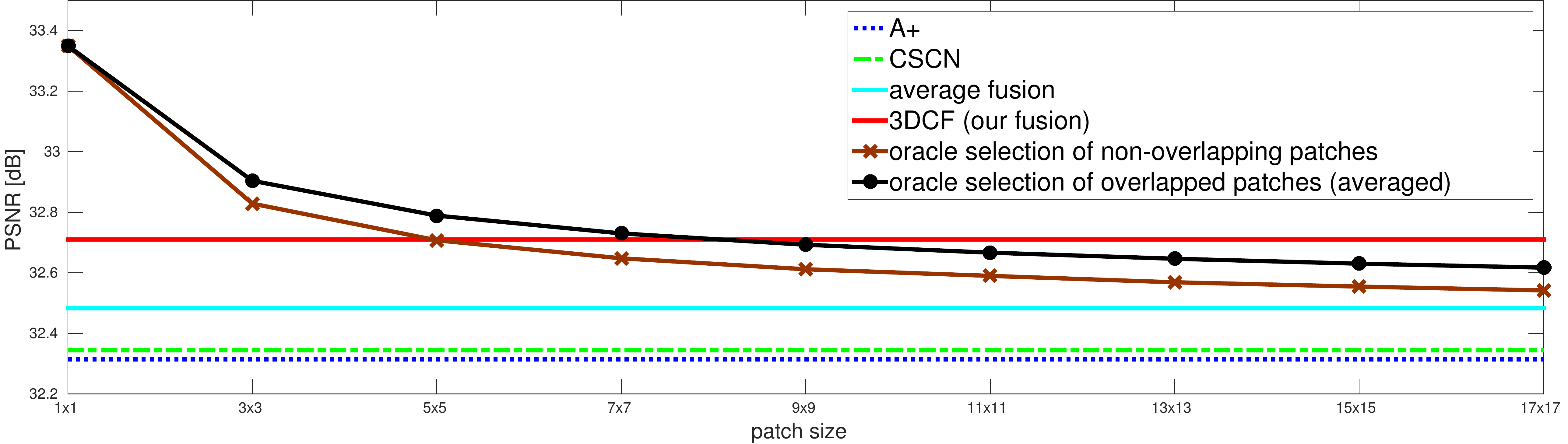}

    \caption{Average PSNR [dB] comparison of A+~\cite{timofte2014a+} and CSCN~\cite{wang2015deep}, average fusion, oracle selection of (overlapping or non-overlapping) patches, and our 3DCF fusion on Set14, upscaling factor $\times2$.}
    \label{fig:fusion_oracle_SR}
  \end{minipage}
\vskip-7pt        
\end{figure}

In Fig.~\ref{fig:fusion_oracle} we report how the chosen patch size affects the performance of a selection strategy, on the same Berkeley images corrupted with AWG noise, $\sigma=50$. We report oracle results, an upper bound for such a strategy. In comparison we report the performance of the fused BM3D and MLP methods, as well as the results of the average fusion and our proposed 3DCF method. We note that i) overlapping patches lead to better results (while significantly slower) than non-overlapping patches; ii) the smaller the patch size the better the oracle results become; iii) the average fusion leads to poorer performance than the fused MLP method; iv) our 3DCF fusion results are comparable with those from the oracle selection strategies for patch sizes above $9\times9$.

Complementary, in Fig.~\ref{fig:fusion_oracle_SR} we start from the A+ and CSCN methods for the super-resolution (SR) task, where we use the Set14 images and an upscaling factor $\times2$ (we use the settings described in the experimental section). As in the denoising case, i) the smaller the patch size is, the better the oracle selection results get; ii) the overlapped patches lead to better fusion results. However, for SR, iii) the average fusion improves over both fused methods; iv) our 3DCF fusion is significantly better than the fused methods, the average fusion, and compares favorably to the oracle selection fusion for patch sizes above $5\times5$.

From these experiments we can conclude that the average and (patch) selection strategies for fusion - while conceptually simple - are either not leading to consistently improved results (case of average fusion) or their oracle upper bounds are quite tight given the difficulty of accurately classifying patches (case of selection strategy). Note that PatchSNR~\cite{Mosseri-ICCP-2013} is an example of a selection strategy and that NN~\cite{Burger-GCPR-2013}, a neural network fusion method, reported better results than PatchSNR.

We therefore followed the combination paradigm for image fusion and design and trained an end-to-end 3D convolutional network from the results of two methods to the targeted restored image.

\section{Learning fine features by 3D convolution}
\label{sec:3D}

\subsection{Motivation and related work}

Most of the existing neural network architectures apply spatial filters which address inputs such as 2D images.
When it comes to videos, thus 3D inputs, these 2D convolutional neural networks (2DCNN) do not employ crucial information such as the temporal correlation.
For example, in human action recognition, the motion information is not captured by 2DCNNs and Ji~\etal~\cite{ji20133d} introduced a 3D convolutional neural network (3DCNN) method.
The 3DCNN architecture has 1 hardwired layer, 3 convolutional layers and 2 subsampling layers. The spatial dimension of inputs $60 \times 40$ are gradually reduced to $1 \times 1$ by going through the network, \ie 7 input frames 
have been converted into a 128-dimensional feature map capturing also the motion information.
In the end, each element of the 128-dimensional feature map is fully connected to each unit in the last layer, then the action class is determined.

For performance improvements a brute force approach that proved successful is to deepen the (neural network) architecture~\cite{wang2015deep,chen2015learning}.
Yet, the improvements decline significantly with the depth while the training time and the demand of hardware (GPU) resources increase.
For example, experiments reported in~\cite{chen2015learning} demonstrate that the bulk of the performance is achieved by the first stages in their denoising TRD method while the last 3 stages (from 8) bring merely 0.01dB to it. In~\cite{Timofte-arxiv-2015} it is shown for SR methods that the first stages are the most important and that adding more stages only slightly improves the performance (of A+) further.

On the other hand, for image restoration tasks such as SR it is common to recover the corrupted luminance component instead of the RGB image directly, and to interpolate the chroma.
However, exploiting the correlation between corrupted RGB or even extra channels such as depth (D) or near-infrared (NIR) should be beneficial to the restoration task at the price of increased computation.
For example, for denoising, Dabov~\etal~\cite{dabov2007color} apply the same grouping method on chroma channels as on the luminance, and they achieve better PSNR performances than by using BM3D~\cite{dabov2007image} independently on three channels.
To sum up, given several highly correlated (corrupted) channels/images, we have a better chance to high quality recovery.

It follows that we can consider the outputs of state-of-the-art methods as highly correlated images, 
which can be treated as the starting point of our proposed novel 3D convolutional fusion (3DCF) architecture.

\begin{figure*}[hpt!]
    \centering
    \includegraphics[height= 5 cm]{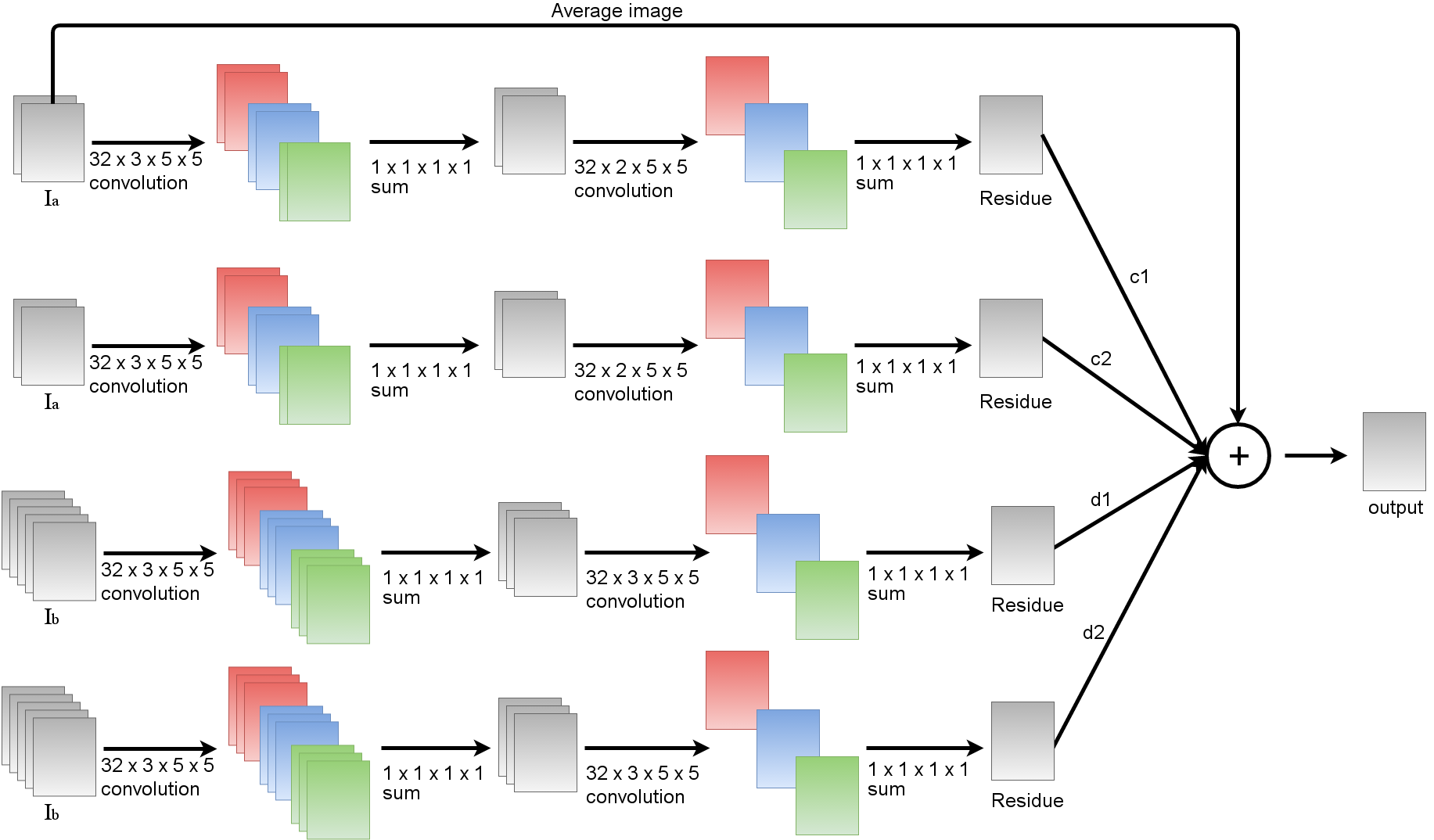}\\

    \caption{\textsf{Proposed 3D convolutional fusion method (3DCF).}}
    \label{fig:3DCF}

\end{figure*}

\subsection{Proposed generic 3D convolutional fusion (3DCF)}
\subsubsection{General Architecture}

As the starting point, we obtain several recovered outputs  $\{ \mathbf{I}_{i} \}_{i = 1, \dots, n}$ from the same corrupted image,  
with different methods.
We stack those highly correlated images along the channel dimension, 
which brings us a multichannel image $\mathbf{I}_{a} = [ \mathbf{I}_{1}, \mathbf{I}_{2}, \dots, \mathbf{I}_{n} ]$ (see Fig.~\ref{fig:3DCF}).

Furthermore, since directional gradient filters are sensitive to intensity changes and edges, and our task is about recovering fine image details based on the results of existing methods, hence the correlation between the recovered output image and its gradients can be exploited.
To this end, we firstly have the naive average input image $\bar{\mathbf{I}} = \frac{1}{n} \sum_{i = 1}^n \mathbf{I}_{i}$, 
then filter it with the first- and second-order gradients, in both the x and y direction, 
\begin{equation}
\label{eq:filter}
\begin{split}
\mathbf{F}_{1x}  &= \begin{bmatrix}
        1 & -1
        \end{bmatrix} =  \mathbf{F}_{1y}^T,\\
\mathbf{F}_{2x}  &= \begin{bmatrix}
        1 & -2 & 1
        \end{bmatrix}/2 =  \mathbf{F}_{2y}^T,  
\end{split}
\end{equation}
followed by stacking those gradient filtered- and average images along the channel dimension,
we have another input $\mathbf{I}_{b}$ as our second starting point,
\begin{equation}
\label{eq:I_b}
\mathbf{I}_b = [ \mathbf{F}_{2x} * \bar{\mathbf{I}},  \mathbf{F}_{1x} * \bar{\mathbf{I}}, \bar{\mathbf{I}}, \mathbf{F}_{1y} * \bar{\mathbf{I}},  \mathbf{F}_{2y} * \bar{\mathbf{I}}].
\end{equation}

Next, we intensively explore the correlation within $\mathbf{I}_{a}, \mathbf{I}_{b}$ by introducing the 3D convolutional layer. 
Related recent works such as ~\cite{chen2015learning,wang2015deep,dong2015image} mainly exploit deep features with spatial filters.
In that case, given the image has multiple channels,
they are independently filtered and eventually summed up as the input for the next layer,
while the correlations among the channels may not be accurately captured.
That is the main reason behind our idea -- to fully explore the fine details along the channel dimension.
As far as we know, this is the first time that a 3D layer is introduced to address low level image tasks.

Our next step is to update the input images $\mathbf{I}_{a,b}$\footnote{Here we abuse of notation,  $\mathbf{I}_{a,b}$ indicates two inputs $\mathbf{I}_{a}, \mathbf{I}_{b}$.} with a 3D hidden layer, 
\begin{equation}
\label{eq:update}
\mathbf{H}_{1}^{a,b}(\mathbf{I}_{a,b}) = \mathop{\text{tanh}} (\mathbf{W}_{1}^{a,b} * \mathbf{I}_{a,b}+ \mathbf{B}_{1}^{a,b}),
\end{equation}
where $\mathbf{W}_1^{a,b}$ correspond $n$ 3D filters with $c \times h \times w$ kernel size and $\mathbf{B}_1^{a,b}$ are biases.
In our design, due to a tradeoff between the memory constraint and speed, 
we recommend $n$ and $c \times h \times w$ to be $32$ and $3 \times 5 \times 5$, rest.
Also we use hyperbolic tangent (tanh) as activation function
because we allow negative value updates to pass through the network rather than ignore them as ReLU~\cite{nair2010rectified} does.
In the following step, we use a naive convolutional layer with a single $1 \times 1 \times 1$ filter, 
which is equivalent to sum up the input $\mathbf{H}_{1}^{a,b}$ 
\begin{equation}
\label{eq:update1}
\mathbf{H}_{2}^{a,b}(\mathbf{H}_{1}^{a,b}(\mathbf{I}_{a,b})) = \mathop{\text{tanh}} (w_{2}^{a,b} \sum_k \mathbf{H}_{1,k}^{a,b}(\mathbf{I}_{a,b})+ b_{2}^{a,b} \mathbf{1}),
\end{equation}
where $w_{2}^{a,b}$, $b_{2}^{a,b}$ are the scalar weights and biases, resp. 
We consider the above two steps as one inference stage.
Another important difference between our proposed method and many other neural network methods is that we reconstruct the image residue instead of the image itself (see Fig.~\ref{fig:3DCF}).
Normally, the perturbation on image residues during the optimization is smaller than the one on image values, which increases the odds that the learning process eventually converges.
Secondly, residue reconstruction substantiates the robust performance of our general architecture for distinct image restoration tasks.
After going through $n$ inference stages, we come to the reconstruction stage, 
\begin{equation}
\label{eq:updaten}
\mathbf{R}_{a,b}(\mathbf{I}_{a,b}) = (w_{2n + 2}^{a,b} \sum  \mathbf{H}_{2n + 1}^{a,b} \circ \mathbf{H}_{2n}^{a,b} \dots \mathbf{H}_{2}^{a,b} \circ \mathbf{H}_{1}^{a,b}(\mathbf{I}_{a,b}) +   b_{2n+2}^{a,b} \mathbf{1}),
\end{equation}
where $\mathbf{R}_{a,b}(\mathbf{I}_{a,b})$ are the image residues we want to predict.
In order to robustify the performance of our network, 
we simply duplicate the above mentioned process for each input image array $\mathbf{I}_a $ and $\mathbf{I}_b$ $n$ times,
which gives us $2n$ separate networks with the same architecture.
In the end we sum up the residues and the average image to obtain our output image $F(\mathbf{I}_1, \mathbf{I}_2, \dots, \mathbf{I}_n)$,
\begin{equation}
\label{eq:sum}
F(\mathbf{I}_1, \mathbf{I}_2, \dots, \mathbf{I}_n) = \frac{1}{n} \sum_k \mathbf{I}_k + \sum_{k}( c_k  \mathbf{R}_{a}^{k}(\mathbf{I}_a) +  d_k  \mathbf{R}_{b}^{k}(\mathbf{I}_b)),
\end{equation}
where $c_k, d_k$ are the coefficients to weight the residues.

\subsubsection{Training}
Our main task is to learn the parameters $\mathbf{\Theta} = (\mathbf{W}, \mathbf{B})$ of the non-linear map $F$.
To this end, we minimize the loss function $l(\mathbf{\Theta})$, 
which computes the Euclidean distance (mean square error (MSE)) between the output image $F(\mathbf{I}_1^i, \mathbf{I}_2^i, \dots, \mathbf{I}_n^i)$ and ground truth image $\mathbf{I}_g^i$ contained in our training set, \ie,
\begin{equation}
\label{eq:loss}
l(\mathbf{\Theta}) = \sum_i \|F(\mathbf{I}_1^i, \mathbf{I}_2^i, \dots, \mathbf{I}_n^i; \mathbf{\Theta}) - \mathbf{I}_g^i \|_2^2.
\end{equation}
The choice of the cost function is appropriate since PSNR is the main evaluation method of image restoration tasks and stands in monotonic relation with MSE.
During the training stage, 
we update the weights/biases with standard back propagation~\cite{rumelhart1988learning,lecun1998gradient}.

Currently, the optimization of the loss function is dominated by the stochastic gradient descent (SGD) method~\cite{bottou2010large}, for example in ~\cite{chen2015learning,wang2015deep,dong2015image}.
Basically, at the $t+1$-th iteration they update the parameters $\mathbf{\Theta}_{t+1}$ with the previous parameter update $\mathbf{\Lambda}_t$ and negative gradient $\nabla l(\mathbf{\Theta})$,  
\begin{equation}
\label{eq:sgd}
\begin{split}
\mathbf{\Lambda}_{t+1}  &= a \mathbf{\Lambda}_{t} - b \nabla l(\mathbf{\Theta}_t),\\
\mathbf{\Theta}_{t+1}  &= \mathbf{\Theta}_{t} + \mathbf{\Lambda}_{t+1},  
\end{split}
\end{equation}
where $a, b$ are the momentum and learning rate, resp.
One weakness of SGD is that the improvements gained from the optimization decrease rapidly with growing iteration steps.
In such case, SGD may not be able to recover accurate details from highly corrupted images.
This is the main reason why we prefer adaptive moment estimation (Adam)~\cite{kingma2014adam} as our optimization method. 
The Adam method is stated as follows,
\begin{equation}
\label{eq:adam1}
\begin{split}
\mathbf{\Lambda}_{t}  &= a_1 \mathbf{\Lambda}_{t-1} + (1 - a_1) \nabla l(\mathbf{\Theta}_t),\\
\mathbf{K}_{t}  &= a_2 \mathbf{K}_{t-1} + (1 - a_2) \nabla l(\mathbf{\Theta}_t)^2,  
\end{split}
\end{equation}
where $a_1, a_2$ are moments and $\mathbf{\Theta}_{t+1}$ is updated based on $\mathbf{\Lambda}_{t}, \mathbf{K}_{t}$,
\begin{equation}
\label{eq:adam2}
\begin{split}
\mathbf{\Theta}_{t+1}  &= \mathbf{\Theta}_{t} - b \frac{\sqrt{1 - (a_2)^t}}{1 - (a_1)^t} \frac{\mathbf{\Lambda}_{t}}{\sqrt{\mathbf{K}_{t}} + \epsilon},
\end{split}
\end{equation}
here $b$ is the learning rate and $\epsilon$ is used to avoid explosion.
At the beginning of the iterations, the cost of $l(\mathbf{\Theta})$ converges considerably faster than SGD. Moreover, Eq.~\eqref{eq:adam2} shows that the magnitudes of parameter updates are independent of the rescaling of the gradient, therefore it provides a relatively fast convergence speed even after a large amount of iterations.  

\section{Experiments}
\label{sec:experiments}

In the following we describe the experimental setup and datasets used to validate our 3DCF approach on both the SR and DN tasks, then discuss the results.

\subsection{Experimental Setup and Datasets}
\label{ssc:datasets}
\subsubsection{DN}
Like most DN-related papers we add white Gaussian (AWG) noise to ground truth images to create our corrupted images.
3 standard deviations $\sigma\in\{ 15, 25, 50\}$ are chosen to measure the performance of 3DCF. 
Under such conditions, we compare our 3DCF with state-of-the-art DN methods as described in the introductory section~\ref{sec:introduction}: BM3D~\cite{Dabov-TIP-2007}, LSSC~\cite{Mairal-ICCV-2009}, EPLL~\cite{Zoran-ICCV-2011}, opt-MRF~\cite{chen2013revisiting},
CRTF~\cite{schmidt2014cascades}, WNNM~\cite{Gu-CVPR-2014}, CSF~\cite{schmidt2014shrinkage}, TRD~\cite{chen2015learning}, MLP~\cite{Burger-CVPR-2012}, as well as the NN~\cite{Burger-GCPR-2013} fusion method.  

We use the same training data mentioned in~\cite{chen2015learning}, 
\ie, 400 cropped images with $180 \times 180$ size from the training part of the Berkeley segmentation dataset (BSD)~\cite{MartinFTM01}.
We evaluate our method on the 68 test images as in~\cite{roth2009fields}, a standard benchmark employed by top methods like~\cite{schmidt2014shrinkage,chen2015learning}.

\subsubsection{SR}
For SR we use the same 3DCF architecture as for DN and test it on the standard benchmarks Set5~\cite{bevilacqua2012low}, Set14~\cite{zeyde2010single} (as proposed in~\cite{timofte2013anchored}) and B100~\cite{timofte2014a+} with 5, 14, 100 images resp.,
which are widely adopted by the recent literature.
To obtain the LR images, according to many of the SR works, we firstly convert the ground truth image into YCbCr color space, then downscale the luminance channel with bicubic interpolation.
Our training data is formed by the 200 training BDS images of size $321 \times 481$ from which we extract millions of LR-HR image pairs.
We report PSNR and SSIM results for the latest methods with top performances: A+~\cite{timofte2014a+}, SRCNN~\cite{dong2015image}, RFL~\cite{schulter2015fast}, SelfEx~\cite{huang2015single}, CSCN~\cite{wang2015deep}.

\subsection{Implementation details}
\label{ssc:implementation_details}
We implement our 3DCF method with Caffe~\cite{jia2014caffe}. 3DCF is used in the same form for both DN and SR.
For clarity and ease of understanding and deployment we prefer stacking two top methods along the channel dimension as our one starting point $\mathbf{I}_a$.
For DN we use MLP~\cite{Burger-CVPR-2012}, an external neural network method, and BM3D~\cite{dabov2007image}, an internal method.
Thus, such combination of two top methods increases our chance to take advantage of the strengths and overcome the weaknesses of both worlds.
For SR, the CSCN~\cite{wang2015deep} and A+~\cite{timofte2014a+} are our favorite because of similar reasons -- one from CNN and another from non-CNN type of methods.
The starting point $\mathbf{I}_b$ is simply obtained by the average image of two methods as well as its corresponding first- and second order gradients along x/y direction.
To enable 3DCF to recover more accurate details, we use two networks for each starting point $\mathbf{I}_a, \mathbf{I}_b$ (See Fig.~\ref{fig:3DCF}), while slightly perturbing the value as the input of each activation, by multiplying $-1$. For the same reason we fix the coefficients $c_1, c_2$ to be 1 and 0.1. So are the coefficients $d_1, d_2$.
Now Eq.~\eqref{eq:sum} looks as follows:
\begin{equation}
\label{eq:sum1}
F(\mathbf{I}_1, \mathbf{I}_2) = \frac{1}{2} (\mathbf{I}_1 +  \mathbf{I}_2) +  \mathbf{R}_{a}^{1}(\mathbf{I}_a) + 0.1\mathbf{R}_{a}^{2}(\mathbf{I}_a) +    \mathbf{R}_{b}^{1}(\mathbf{I}_b) + 0.1\mathbf{R}_{b}^{2}(\mathbf{I}_b).
\end{equation}
For the sake of time complexity and memory saving, each network showed in Fig.~\ref{fig:3DCF} has 4 layers, and the filter size $n \times c \times h \times w$ is set to be $(32 \times 3 \times 5 \times 5, 1 \times 1 \times 1 \times 1, 32 \times 3 \times 5 \times 5, 1 \times 1 \times 1 \times 1)$ for $\mathbf{I}_a, \mathbf{I}_b$.
We also set the channel-, height- and width stride to be 1 for all layers.
It is expected that our output is a single image with the same spatial size as the input image.
To this end, the channel-, height- and width padding size are determined to be $(1 \times 2 \times 2, 0 \times 0 \times 0, 0 \times 2 \times 2, 0 \times 0 \times 0)$ for $\mathbf{I}_a$, and for $\mathbf{I}_b$ we follow the same setup except the first layer parameters are determined to be $0 \times 2 \times 2$.  
We also initialize the weights by a Gaussian distribution with standard deviation 0.05 for convolutional layers, and put the weight to 1 for sum layers, and the bias to 0 for all cases.

Meanwhile, we simply use the default learning- and decay rate 1 when learning the weights/biases for each layer. 
In the end, for Eq.~\ref{eq:adam2} the learning rate $b$ for the whole network is considered to be 0.001, the moments $a_1, a_2$ have the default value 0.9, 0.999, and $\epsilon$ is also set to the default $10^{-8}$.
It is worth mentioning that all the parameters are exactly the same for the two tasks, DN and SR.

\subsection{Denoising results}
\label{ssc:denoising_results}
We demonstrate our 3DCF method on 68 standard images~\cite{roth2009fields} from BSD~\cite{MartinFTM01}.
We apply the best setup for the compared methods, already described in the introductory section~\ref{sec:introduction}.
CRTF~\cite{schmidt2014cascades} has 5 cascades, CSF~\cite{schmidt2014shrinkage} employs the $7 \times 7$ filter, the same as TRD~\cite{chen2015learning} with 8 stages.
Table~\ref{tb:dn} shows that our 3DCF method achieves top performances compared to other methods for 3 different standard deviations.
For example, if we start our method with BM3D~\cite{Dabov-TIP-2007} and MLP~\cite{Burger-CVPR-2012},
we are 0.11dB and 0.1dB better than the top standalone method MLP for $\sigma\in\{ 25, 50\}$.
Due to the lack of an MLP model trained for $\sigma = 15$, 
we use BM3D+TRD instead.
Still, the performance of our 3DCF is consistent with the other cases, 0.09dB higher than TRD, the currently best method.
Interestingly, if we compare 3DCF with the NN fusion method under the same conditions, that is, with the same starting methods BM3D and MLP,
the proposed method outperforms NN with 0.15 and 0.07dB for $\sigma\in\{ 15, 25\}$.
Such observation confirms the non-trivial improvements achieved by 3DCF.
Moreover, Fig.~\ref{fig:dn1} indicates that the naive average of MLP and BM3D is even worse than MLP. 
Besides, it is also notable from Fig~\ref{fig:dn1} that the PSNR gradually increases with the growth of back propagation.
3DCF is robust to the fused methods, TRD + MLP leads to relative improvements comparable with those achieved starting from BM3D+MLP or BM3D+TRD.

\begin{table}[]
\centering

  \begin{minipage}[c]{0.4\textwidth}

\caption{Average PSNR values [dB] on 68 images from BSD dataset as in~\cite{roth2009fields} for $\sigma \in \{ 15, 25, 50\}$. The best is with bold.  
The results with (*) are obtained from~\cite{chen2015learning}.}

        \label{tb:dn}
        \end{minipage}\hfill
  \begin{minipage}[c]{0.55\textwidth}
  \centering
  \resizebox{0.65\textwidth}{!}
  {
  \centering
  \begin{tabular}{l|ccc}
\multirow{2}{*}{Method} & \multicolumn{3}{c}{$\sigma$} \\ 
                                & 15      & 25      & 50     \\ \hline
BM3D~\cite{Dabov-TIP-2007}      & 31.08   & 28.57   & 25.61  \\ 
*LSSC~\cite{Mairal-ICCV-2009}    & 31.27   & 28.70   & 25.72  \\ 
*EPLL~\cite{Zoran-ICCV-2011} & 31.19   & 28.68   & 25.67  \\ 
*opt-MRF~\cite{chen2013revisiting}                 & 31.18   & 28.66   & 25.70  \\ 
*$\mathop{\text{CRTF}}_5$~\cite{schmidt2014cascades}                 &         & 28.75   &        \\ 
*WNNM~\cite{Gu-CVPR-2014}                    & 31.37   & 28.83   & 25.83  \\ 
$\mathop{\text{CSF}}_{7 \times 7}$~\cite{schmidt2014shrinkage}
                        & 31.24   & 28.71   &        \\ 
$\mathop{\text{TRD}}_{7 \times 7}^8$~\cite{chen2015learning}
                        & 31.42   & 28.93   & 25.99  \\ 
MLP~\cite{Burger-CVPR-2012}                     &         & 28.96   & 26.01   \\ 
NN (BM3D+MLP)~\cite{Burger-GCPR-2013}           &         & 28.92   & 26.04   \\ \hline
3DCF (BM3D+TRD)               & \textbf{31.51}   &  29.03       & 26.10       \\
3DCF (BM3D+MLP)                &         & \textbf{29.07} & 26.11       \\ 
3DCF (TRD+MLP)                &         & \textbf{29.07}        & \textbf{26.12}       \\ 
\end{tabular}
}

  \end{minipage}
\vskip-14pt        
\end{table}

\begin{figure*}
  \centering
\includegraphics[width=1.0\textwidth]{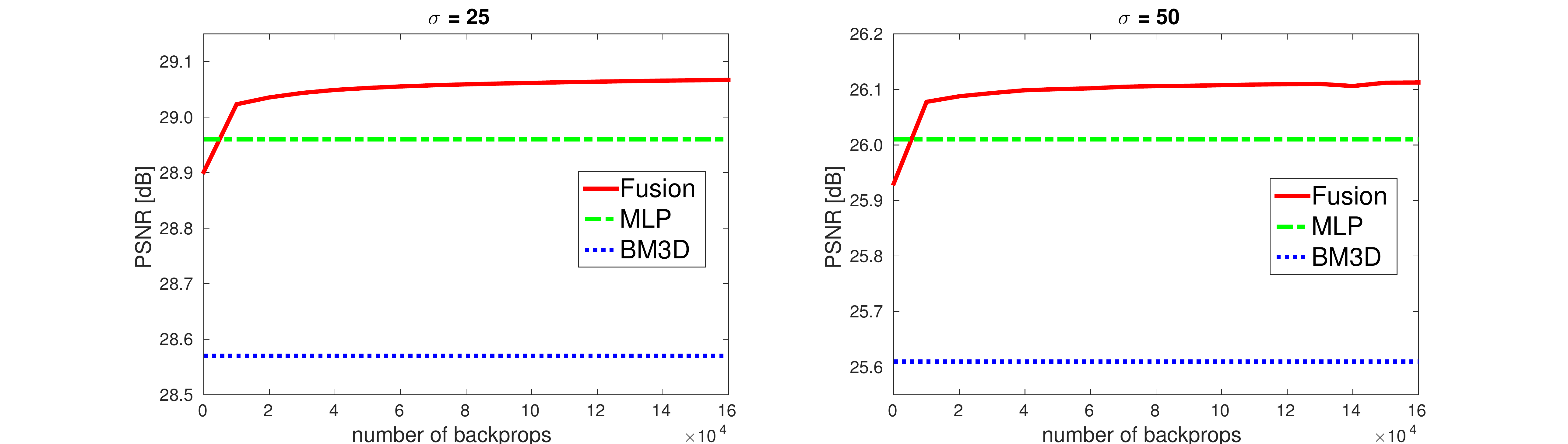}\\
\caption{PSNR versus backprops on 68 images for $\sigma\in\{25, 50\}$.}
\label{fig:dn1}
\end{figure*}

\subsection{Super resolution results}
\label{ssc:SR_results}
The PSNR and SSIM results are listed in Table~\ref{tb:sr}.
Here our 3DCF fuses A+~\cite{timofte2014a+} with CSCN~\cite{wang2015deep}. 
Note that we modify the steps of downscaling the image for CSCN to be consistent with other methods including A+ and SRCNN. That is the reason why we obtain different PSNR results for CSCN than in the original work~\cite{wang2015deep}. 
As in the case of DN, our 3DCF shows significant improvements over the starting methods.
The PSNR improvements vary from 0.11dB on (B100, $\times3$) to 0.35dB on (Set 5,$\times2$) over the best result from SRCNN(with largest model). The SSIM improvements follow the same trend.
Note that for SR, the naive average fusion of A+ and CSCN results improves over both fused methods. However, our 3DCF results are on average 0.2dB higher than the average fusion, as shown in Fig.~\ref{fig:sr1}.

\begin{table}[thp!]

\caption{Average PSNR/SSIMs for upscaling factors $\times2$, $\times3$, and $\times4$ on datasets Set5, Set14, and B100. The best is with bold.}

\resizebox{\linewidth}{!}{
\setlength{\tabcolsep}{0.5em}
\begin{tabular}{cccccccc}
\hline
\multirow{2}{*}{Dataset} & \multirow{2}{*}{Scale} & A+~\cite{timofte2014a+}        & SRCNN~\cite{dong2015image}     & RFL~\cite{schulter2015fast}      & SelfEx~\cite{huang2015single}       & CSCN~\cite{wang2015deep}    & \textbf{3DCF (CSCN+A+)} \\ \cline{3-8} 
                         &                        & PSNR/SSIM & PSNR/SSIM & PSNR/SSIM & PSNR/SSIM & PSNR/SSIM & PSNR/SSIM \\ \hline
\multirow{3}{*}{Set 5}   & x2                     &36.56/0.9612&36.68/0.9609&36.52/0.9589&36.50/0.9577           &36.55/0.9605           &\textbf{37.03}/\textbf{0.9631}           \\ 
                         & x3                     &32.67/0.9199&32.83/0.9198&32.50/0.9164&32.63/0.9190           &32.68/0.9197           &\textbf{33.11}/\textbf{0.9255}           \\ 
                         & x4                     &30.33/0.8749&30.52/0.8774&30.17/0.8715&30.32/0.8728           &30.44/0.8779           &\textbf{30.82}/\textbf{0.8865}           \\ \hline
\multirow{3}{*}{Set 14}  & x2                     &32.32/0.9607&32.52/0.9612&32.30/0.9599&32.27/0.9584           &32.36/0.9593           &\textbf{32.71}/\textbf{0.9623}           \\ 
                         & x3                     &29.16/0.8869&29.35/0.8886&29.07/0.8842&29.19/0.8873           &29.19/0.8850           &\textbf{29.48}/\textbf{0.8907}           \\ 
                         & x4                     &27.33/0.8277&27.53/0.8285&27.23/0.8251&27.43/0.8279           &27.41/0.8256           &\textbf{27.69}/\textbf{0.8334}           \\ \hline
\multirow{3}{*}{B100}    & x2                     &31.16/0.8857&31.32/0.8874&31.13/0.8842&31.15/0.8860                                    &31.20/0.8836           &\textbf{31.48}/\textbf{0.8899}           \\ 
                         & x3                     &28.25/0.7824&28.37/0.7853&28.20/0.7814&28.25/0.7821           &28.28/0.7804           &\textbf{28.48}/\textbf{0.7881}           \\ 
                         & x4                     &26.76/0.7073&26.86/0.7089&26.70/0.7068&26.81/0.7078           &26.83/0.7072           &\textbf{26.99}/\textbf{0.7147}           \\ \hline
\end{tabular}
}

\label{tb:sr}
\end{table}

\begin{figure*}
  \centering

\includegraphics[width=1.0\textwidth]{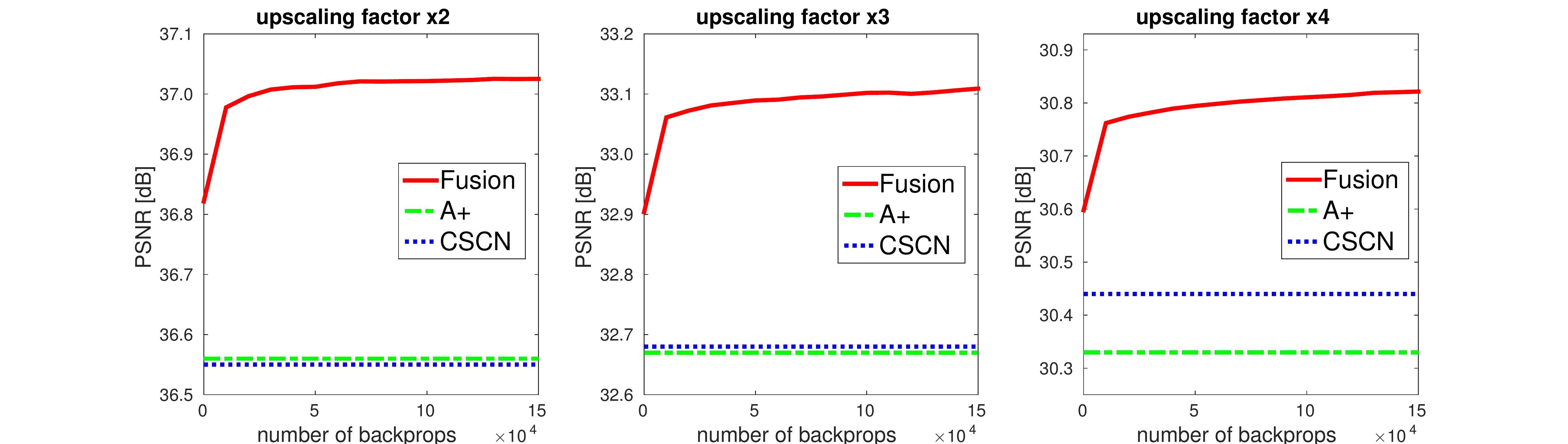}\\

\caption{PSNR versus Iterations on Set 5 dataset for upscaling factors $\times2$, $\times3$, $\times4$.}
\label{fig:sr1}

\end{figure*}

\subsection{Other aspects}
\label{ssc:discussion}

\noindent{\bf Running time }
3DCF runs on roughly 0.04 second per $321\times480$ image on nVidia TitanX GPU, which is quite competitive and shows that at the price of slight increase in processing time one could fuse available image restoration results.
3DCF needs about 5 hours training time to obtain meaningful improvements over the fused methods, and this is mainly due to the Adam method.

\noindent{\bf General }
To summarize, our 3DCF method shows wide adaptability for two important image restoration tasks, DN and SR, with non-trivial improvements.
Also, the training and running times of 3DCF are competitive in comparison with other neural network architectures.
We must admit that for certain combinations of existing methods our proposed fusion method only shows mild progress, for example for the case of TRD+MLP (see Table~\ref{tb:dn}). 
The sensitivity to the starting point drives us to be careful of the choice of starting methods. 

\section{Conclusions}
\label{sec:conclusions}
We propose a novel 3D convolutional fusion (3DCF) network for image restoration. With the same settings, for both single image super resolution and image denoising, we achieve significant improvements over the fused methods and other fusion methods on several standard benchmarks. For speeding up the training, we apply an adaptive moment estimation method.
The testing and training times are also competitive to other recent deep neural networks.

\bibliography{camera_ready}
\bibliographystyle{plain}

\end{document}